\documentclass[11pt]{article}

\usepackage[final]{acl}

\usepackage{times}
\usepackage{latexsym}
\usepackage[T1]{fontenc}
\usepackage[utf8]{inputenc}
\usepackage{microtype}
\usepackage{inconsolata}
\usepackage{graphicx}
\usepackage{booktabs}
\usepackage{amsmath}
\usepackage{amssymb}
\usepackage{multirow}
\usepackage{xcolor}
\usepackage{cleveref}

\newcommand{\modelname}{Qwen3-0.6B}

\title{Hybrid Generative--Tagging Decoding for Efficient\\ Product Attribute--Value Extraction}

\author{Keshav Kolluru \hskip 1em Manas Gupta \hskip 1em Maroof Ahmad \hskip 1em Suchitra Sathyanarayana \\ 
  Amazon \\
  \texttt{\{keshavkl,manasgup,maroahma,suchisat\}}@\texttt{amazon.com} \hskip 1em \\}

\begin{document}
\maketitle

\begin{abstract}
Attribute--value extraction is important for building high-quality product catalogs in e-commerce. 
Existing systems either rely on generative or tagging strategies for extracting the attribute names and their corresponding values. 
Tagging models directly label spans in the source, making them fast and grounded, but they cannot produce values that require reformatting or generate rare attribute names outside their label space. 
Generative models avoid these drawbacks by generating both attribute names and values token by token, but this auto-regressive nature makes them slow and limits attribute coverage. 
We introduce \emph{hybrid generative-tagging decoding}, a novel model architecture that combines tagging source spans and open-ended generation.
For frequent attributes whose values are copyable, a tagging head extracts values directly from the source.
This lets the generative head focus on the remainder -- rare attributes outside the tagger's label space and non-copyable values, including implicit or normalized values.
Integrated into vLLM inference, the hybrid decoder raises attribute coverage by 10\% and extracts 63\% more correct attribute--value pairs at no precision cost, while reducing latency by 44\%.
\end{abstract}

\section{Introduction}

Structured attribute–value pairs like (\texttt{brand:}~Nike, \texttt{color:}~red) power foundational e-commerce catalog operations like product matching~\citep{primpeli2019wdc} and serve as the critical knowledge base for modern agentic web navigators~\citep{yao2022webshop} and conversational shopping assistants~\citep{lyu2025deepshop}.
Merchant-provided catalog fields on product pages are often noisy, and incomplete, with key attributes missing altogether~\citep{yang-etal-2022-mave,subhalingam2025framework}.
Manual cleanup does not scale, making automatic, high-throughput attribute extraction essential for keeping fast-moving catalogs accurate enough for downstream shopping systems~\citep{yan-etal-2021-adatag,zhang-etal-2024-stronger}.

\begin{table}[t]
\centering\small
\setlength{\tabcolsep}{4pt}
\caption{A product listing with both attribute types. \emph{Taggable} values are contiguous spans of the source and can be copied verbatim; \emph{generative} values must be reformatted or are only implied by the text.}
\label{tab:examples}
\begin{tabular}{@{}p{0.32\columnwidth}p{0.60\columnwidth}@{}}
\toprule
\multicolumn{2}{@{}p{\columnwidth}@{}}{\centering \textbf{Product:} \emph{``Mrs Ball's Original Chutney. Ingredients: Sugar, Vinegar, Water, Peaches, Apricots, Modified Maize Starch, Spices. Imported from South Africa.''}\par} \\
\midrule
\multicolumn{2}{@{}l}{\textbf{Taggable attributes} \emph{(label source span)}} \\
\texttt{brand} & Mrs Ball's \\
\texttt{product\_name} & Original Chutney \\
\texttt{ingredients} & Sugar, Vinegar, Water, Peaches, Apricots, Modified Maize Starch, Spices \\
\midrule
\multicolumn{2}{@{}l}{\textbf{Generative attributes} \emph{(reformat / implicit)}} \\
\texttt{product\_type} & condiment \\
\texttt{cuisine} & South African \\
\texttt{diet\_type} & vegetarian \\
\bottomrule
\end{tabular}
\end{table}

Existing Attribute-Value Extraction (AVE) systems typically choose one of two regimes with complementary strengths. 
Tagging models label source spans with attribute names~\citep{yan-etal-2021-adatag,yang-etal-2022-mave}. 
They are fast and grounded in the input, but they can only return values that appear as spans and generally operate over a fixed label space.
Generative LLMs instead produce attribute names and values as text~\citep{khandelwal-etal-2023-large,zou-etal-2024-eiven,li-etal-2025-open}. 
This supports normalisation, implicit values, and rare attributes, but every value must be decoded autoregressively, making inference expensive at catalog scale. 
Moreover, a generative model trained on incomplete annotations learns to emit about as many attributes as it saw labeled, which limits its coverage.
  
Product pages routinely require both capabilities at once. 
The listing in \Cref{tab:examples} pairs values that appear verbatim and should be copied exactly (\texttt{brand}, \texttt{product\_name}, \texttt{ingredients}) with values that must be generated: a \texttt{product\_type} (condiment) implied by the description, a \texttt{cuisine} (South African) read off `Imported from South Africa', a \texttt{diet\_type} (vegetarian) inferred from the ingredients. 
We call the first class \textbf{taggable} and the second \textbf{generative}. 
In our corpus 81.1\% of values are taggable and 18.9\% generative, and both co-occur within most records.

We propose \emph{hybrid generative-tagging decoding}, a single-model strategy that splits attribute extraction into two channels that share a single LLM prefill over the product text.
A tagging channel reads the prompt hidden states and scores the $K{=}500$ most frequent attribute names, which cover 87.8\% of attributes in our corpus; detected attributes are filled by copying contiguous source spans verbatim, without any decoding.
The generative channel is an autoregressive decoder that emits only the \emph{remainder}: attributes whose values are implicit/normalized, attributes outside the frequent-attribute vocabulary, and in-vocabulary attributes whose values are not contiguous source spans.
The final output merges the attribute-value pairs produced from both the channels.

The tagging channel uses a lightweight \emph{span head}.
Each vocabulary attribute is represented by a query, the LM's own encoding of the attribute name.
The head \emph{locates} an attribute by scoring every input position as a possible value start against a learned \texttt{NULL} option that lets it abstain, and \emph{delimits} the value with a second pointer that scores end positions; start and end are independent pointers into the same prefill states.
All $K$ attributes are resolved in one batched pass over the prefill states, so copied values add grounded content without any extra decoding steps.
Since each attribute is tested independently, the span head can surface more attributes than present in the incomplete training set.

Our contributions are:
\begin{itemize}
\item We introduce \textbf{hybrid generative-tagging decoding}, which shares a single LLM prefill between an autoregressive generative channel and a tagging channel, so that copyable values are extracted with no decoding steps while generation is reserved for implicit values.
\item We design a \textbf{span head} that finds the spans for $K$ frequent attributes with a learned \texttt{NULL} slot to detect attribute presence.
\item We show that hybrid decoding is both \textbf{faster and higher-recall} than a generative baseline: it raises attribute coverage from 69.8\% to 80.2\% and on average extracts 33.8 correct attribute--value pairs against 20.7 pairs from baseline, with precision maintained at 97\% in both cases. 
The mean response time reduces from 56.6 ms to 31.5 ms, a \textbf{1.80$\times$} speed-up.

\end{itemize}

\section{Related Work}
\label{sec:related}

E-commerce attribute--value extraction (AVE) turns noisy, free-form product text into structured (attribute, value) pairs, a long-standing goal of web information extraction~\citep{cohen-2001-issues,mecca-etal-2001-roadrunner}. 
It is studied over large product corpora and benchmarks~\citep{logan-etal-2017-mae,yang-etal-2022-mave,primpeli2019wdc} and increasingly exploits signals beyond raw text, such as, DOM structure, layout, and images~\citep{zhu-etal-2020-multimodal,zhou-etal-2022-lantern,wang-etal-2023-mustie,deng-etal-2022-dom-lm,li-etal-2022-markuplm,farag-etal-2025-pinterest}. 
Most systems fall under one of two regimes, generation or tagging, with complementary strengths. 

\paragraph{Generative extraction.} Generative approaches cast AVE as conditional text generation, decoding attribute names and values directly with sequence-to-sequence or LLM backbones. 
This framing scales naturally to open label spaces and has been pushed toward large-scale~\citep{khandelwal-etal-2023-large}, implicit~\citep{zou-etal-2024-eiven}, open-world~\citep{li-etal-2025-open}, visual~\citep{gong-etal-2025-visual,gong-etal-2025-mice}, and lifelong~\citep{zhang-etal-2024-stronger} settings. 
Because values are emitted as free text, generation handles normalization, implicit values, and rare attributes; but it decodes every value autoregressively, making inference costly at catalog scale.

\paragraph{Tagging extraction.} Tagging approaches label spans of the product text with attribute names, inheriting the long line of sequence-labeling and span-chunking methods in NLP~\citep{ramshaw1995text,lafferty-etal-2001-crf}. 
For AVE, OpenTag \citep{zheng-etal-2018-opentag} introduced neural sequence tagging for open-world value extraction, while SUOpenTag \citep{xu-etal-2019-scaling} scaled this formulation to thousands of attributes by treating the attribute as a query and sharing a global BIO tag set. 
AdaTag~\citep{yan-etal-2021-adatag} uses attribute-conditioned decoding to extract values as source spans, MAVE~\citep{yang-etal-2022-mave} supplies large-scale span supervision over Amazon data, and joint formulations predict attributes and extract their values together~\citep{zhu-etal-2020-multimodal}. 
Closely related QA formulations treat each attribute as a question and identify its answer span using an encoder shared across attributes~\citep{wang-etal-2020-learning}.
EAVE improves the efficiency of repeated extraction across many attributes by caching product-context representations and reusing them through lightweight attribute--context interactions~\citep{yang-etal-2024-eave}.
Tagging is fast but it can only returns values that appear contiguously in the text and cannot extract attributes outside its vocabulary.

GenToC~\citep{subhalingam2025framework} also mixes generation and tagging paradigms, but only for ability to learn from noisy data: a generator proposes candidate attributes that a token-classification model then fills with values. 
We instead run both channels \emph{concurrently} for the purpose of efficiency.
  
\paragraph{Efficient inference.} 
General-purpose serving systems raise throughput through paged memory management and structured  execution~\citep{kwon2023vllm,zheng2024sglang}, while assisted decoding such as speculative decoding cuts the latency of the autoregressive loop~\citep{leviathan2023speculative}. 
These methods take the number of generated tokens as given and make each token, or each request, cheaper. 
Our hybrid decoder is orthogonal to these methods: it reduces the number of tokens that must be decoded autoregressively in the first place, so its speed-ups stack on top of the vLLM serving stack.
  
\section{Method}
\label{sec:method}

\begin{figure*}[t]
\centering
\includegraphics[width=\textwidth]{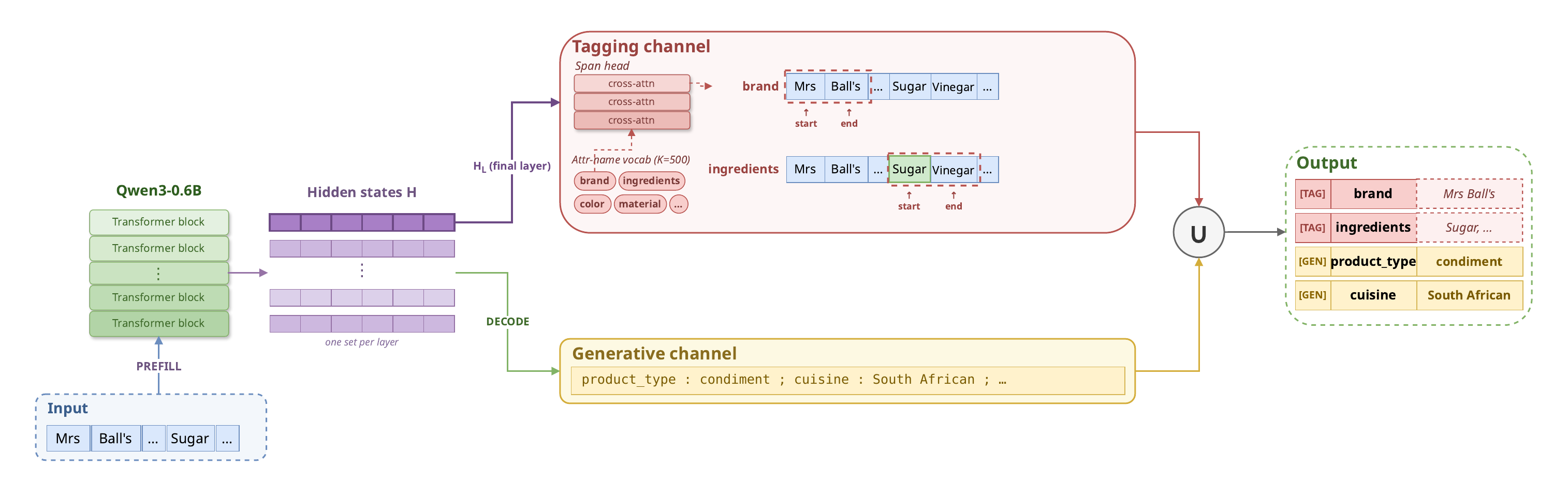}
\caption{Hybrid decoder with two decoupled channels over a shared prefill. During prefill the backbone maps the source tokens to hidden states $H$. The \emph{generative channel} (\texttt{[GEN]}) autoregressively decodes only the remainder---generative, out-of-vocabulary, and non-copyable attributes---as inline \texttt{name : value} pairs. In parallel, the \emph{tagging channel} (\texttt{[TAG]}) runs a span head over $H$ for all $K$ vocabulary attributes at once, claiming an attribute against a learned per-attribute \texttt{NULL} score and copying the chosen source span verbatim. The two channels are merged by attribute name, preferring the generative value on collision.}
\label{fig:arch}
\end{figure*}

The hybrid decoder uses a per-attribute routing decision: copy a value when it is a verbatim source span, generate it otherwise. 
A product $p$ is processed once during prefill into prompt hidden states $H=(h_1,\dots,h_L)\in\mathbb{R}^{L\times d}$, which feed two channels that read $H$ (\Cref{fig:arch}): a generative channel and a tagging channel.
An in-vocabulary attribute is tagged when its value is a verbatim source span and generated otherwise. 
We first describe the two channels, then the span head that powers the tagging channel (\Cref{sec:head}), and the training and inference of the model (\Cref{sec:objective}).

\paragraph{Tagging channel.} A lightweight span head scores a fixed vocabulary $\mathcal{V}$ against the last layer hidden states $h_L$. We set $\mathcal{V}$ to the $K{=}500$ most frequent attribute names, which cover $87.8\%$ of attribute instances in our corpus.
For each $k\in\mathcal{V}$, the head either abstains with \texttt{NULL} or returns a source span $(s,e)$, which becomes the value of attribute $k$.
All $K$ attributes are scored in one batched pass over the prefill states, ensuring efficient computation. 

\paragraph{Generative channel.} The model autoregressively decodes a semicolon-delimited list of \texttt{name}: value; pairs. This channel is responsible for the \emph{remainder}: attributes outside the tagging vocabulary, attributes whose values are implicit or require reformatting, and in-vocabulary attributes where value is not present as a contiguous span in the input.

\subsection{Span head}
\label{sec:head}

The span head is shared across all $K$ attributes: a cross-attention block followed by two bilinear pointers and a learned null scorer. For each attribute $k\in\mathcal{V}$, it first produces a contextualized query $q^{(1)}_k$, then factorizes extraction into two decisions: where the value begins and ends. The start decision also scores a learned \texttt{NULL} option, to determine presence of the attribute.

\paragraph{Attribute queries.} We represent attribute $k$ with the base model's own encoding of the attribute-name string. 
We take $q_k$ to be the mean-pooled final hidden state of that string. 
During training, these name encodings are computed in-batch and updated jointly with the base model. 
At inference, the fine-tuned model encodes all $K$ names once into a cached table $Q\in\mathbb{R}^{K\times d}$.

\paragraph{Locate and claim.} A cross-attention block lets the attribute query attend to all prompt positions,
\[
q^{(1)}_k=\mathrm{CrossAttn}(q_k,H).
\]
The head therefore obtains document-level evidence for existence and boundary selection while reusing hidden states that the model has already computed. 
It then scores every start position and one learned \texttt{NULL} slot,
\[
s_{k,i}=(q^{(1)}_k)^\top W_s\,h_i,\qquad
\nu_k=g_\varnothing(q^{(1)}_k,q_k),
\]
where $W_s\in\mathbb{R}^{d\times d}$ and $g_\varnothing$ is a learned null scorer. The $L$ start logits and $\nu_k$ are normalized together as an $(L{+}1)$-way softmax, making \texttt{NULL} the model's abstention option for attribute $k$ on product $p$. The attribute is \emph{claimed} if its best start outscores \texttt{NULL}.

\paragraph{Delimit.} From the same contextualized query $q^{(1)}_k$, a second bilinear pointer with $W_e\in\mathbb{R}^{d\times d}$ scores every end position in the prompt,
\[
e_{k,j}=(q^{(1)}_k)^\top W_e\,h_j,\qquad j\in[1,L].
\]
Start and end are independent pointers into the same hidden states, so the span score factorizes as $P(s)\,P(e)$. At decode time, the end is restricted to a forward window $j\in[s,s{+}B)$ from the chosen start. This cap limits maximum span length and ties the end to the same occurrence as the start, which is important on product pages where values and catalog fields often repeat. The independent end pointer is more accurate than an end head anchored on the chosen start token (\Cref{sec:experiments}).

\subsection{Training and Inference}
\label{sec:objective}

\paragraph{Route supervision.} Training targets follow directly from the routing rule. For every in-vocabulary gold pair, we search the source for contiguous token-exact occurrences of the gold value. If at least one occurrence exists, the pair is routed to the tagging channel and removed from the generative target; otherwise it remains in the generative target. Out-of-vocabulary pairs always stay generative. This
defines the two targets that the losses below are computed against.

The model is trained end to end with
\[
\mathcal{L}=\mathcal{L}_{\text{gen}}+\mathcal{L}_{\text{tag}},
\]
where $\mathcal{L}_{\text{gen}}$ is standard causal-LM cross-entropy over the generative remainder and $\mathcal{L}_{\text{tag}}$ supervises the span head from span-level occurrence sets. 

\paragraph{Positive span loss.} For a locatable gold pair, let $\mathcal{O}_k(p)$ be the set of all token-exact occurrences $(s,e)$ of attribute $k$'s value in product $p$. 
Since the same value may appear multiple times, we treat every occurrence as valid and marginalize over them:
\[
\mathcal{L}_{\text{span}}(k,x)=-\log\!\!\sum_{(s,e)\in\mathcal{O}_k(x)}\!\!P_k(s)\,P_k(e).
\]
Here $P_k(s)$ is the start probability from the $(L{+}1)$-way softmax and $P_k(e)$ is the end probability from an independent softmax over prompt positions. Each summand scores a complete occurrence with matched boundaries, avoiding credit for combining the start of one mention with the end of another.

\paragraph{Negative \texttt{NULL} loss.} For attribute--record pairs that should not be extracted by the tagging channel, we maximize the \texttt{NULL} probability,
\[
\mathcal{L}_{\text{null}}(k,x)=-\log P_k(\texttt{NULL}).
\]
We sample 30 negatives per example, including hard negatives whose attribute is in vocabulary but whose value is not locatable and therefore belongs to the generative channel. 
These cases teach the head to abstain even when an attribute is semantically relevant to the page but should not be copied as a clean span.
For a minibatch, $\mathcal{L}_{\text{tag}}$ is the average of $\mathcal{L}_{\text{span}}$ over locatable in-vocabulary pairs and $\mathcal{L}_{\text{null}}$ over sampled negative attribute--record pairs.

\paragraph{Inference.}
\label{sec:infer}
At inference, a single prefill serves both channels. 
The span head scores all $K$ vocabulary attributes in one batched pass, while the base model decodes the generative remainder. 
For every claimed attribute, we choose the argmax start position and then the argmax end position within $[s,s{+}B)$; the enclosed source tokens are copied verbatim as the value.
Since the head is non-autoregressive and operates on prefill states, span resolution requires no additional decoding. 
The final prediction is the union of the two channels, on collision we keep the generative value for the attribute.

\section{Experimental Setup}
\label{sec:experiments_setup}

\paragraph{Dataset.} 
We use a corpus of 111K randomly sampled product records from Amazon Catalog of US marketplace. 
Each record pairs all available catalog text with attribute--value annotations produced by Kimi-K2.5~\citep{kimiteam2026kimik25visualagentic} and labeled as taggable or generative attributes. 
The corpus contains 2.62M attribute--value pairs; 81.1\% are taggable, 18.9\% are generative, the split that motivates routing copyable values to the tagging channel. 
We restrict this paper to text-sourced attributes, train on 100{,}195 records, and evaluate on a held-out validation set of 11{,}132 records.
\Cref{tab:data} summarizes the resulting data. 
We additionally evaluate on two public attribute--value extraction datasets: AE-110K~\citep{xu-etal-2019-scaling} and OA-Mine~\citep{zhang2022oamine}, using the train/test splits released by \citet{brinkmann2025extractgpt}.

\paragraph{Metrics.}
A model can express a correct attribute-value in a form that differs from the ground truth, because of normalization, paraphrasing, or formatting differences.
An exact match between prediction and the ground-truth pair would wrongly penalize cases that are in fact correct.
We therefore use an LLM judge: we show GPT-5.6 Luna the source text and a system's emitted pairs, and it labels each pair as correct, incorrect, or unverifiable against the source, allowing reasonable normalizations.
Over a manual audit on 4,795 pair attribute-value pairs, we find that human judgment agrees with the LLM judge on 99.04\% of pairs ($\kappa = 0.867$).
\textit{Precision} is measured over all emitted pairs for correctness, as evaluated by the LLM judge. Unverifiable pairs are rare ($<0.34\%$) and are treated as incorrect.
\textit{Recall} measures the fraction of a product's attributes the system recovers: we report the fraction of ground-truth attributes that get an exact matching prediction.
\textit{Num. Extractions} is the average number of pairs a system emits per product, an indicator of recall independent of the ground truth.
\textit{Quality} is a blinded side-by-side comparison done by human annotators: for each record an annotator sees all three systems' outputs in random, anonymized order and rates each from $1$ to $5$ for overall correctness, completeness, and grounding; we report the mean and variance over $1{,}000$ records.
\textit{Speed} is the average per-product wall-clock time under identical batched processing.

\paragraph{Training.} 
We use \modelname{} as the base LLM for all the models, unless stated otherwise.
We fine-tune models with bf16 mixed precision and DeepSpeed ZeRO-2 on 8 A100 GPUs for 3 epochs, using learning rate $2\times10^{-5}$ and maximum sequence length 4096. The span head uses band width $B{=}96$ and up to 30 sampled negatives per example.
For inference, we use vLLM~\citep{kwon2023vllm} and a single A100 GPU. 

\begin{table}[t]
\centering\small
\setlength{\tabcolsep}{4pt}
\caption{Statistics of the corpus. Both splits average 22 attribute-value pairs per product, of which about 88\% are taggable and 12\% generative.}
\label{tab:data}
\begin{tabular}{lrrrrr}
\toprule
\textbf{Split} & \textbf{\#Products} & \textbf{Avg. AVE} & \textbf{Extr\%} & \textbf{Gen\%} \\
\midrule
Train      & 100{,}195 & 22.3 & 88.3 & 11.7 \\
Validation & 11{,}132  & 22.4 & 88.2 & 11.8 \\
\bottomrule
\end{tabular}
\end{table}

\section{Experiments}
\label{sec:experiments}

In this section, we compare the performance of different systems on both internal and public datasets.
The \textbf{Hybrid} model uses the method in \Cref{sec:method}, with the $K{=}500$ attribute vocabulary. \Cref{app:ksweep} sweeps $K$ and motivates this choice.
The \textbf{Generative} model decodes every attribute--value pair autoregressively as \texttt{name}: value.
The \textbf{Tagging} model is the hybrid's tagging channel served on its own (the same span head and base model, with generative decoding disabled), so it isolates the cost and grounding of tagging under an otherwise identical setup.
We also compare with recently published attribute-value extraction systems: \textbf{GenToC}~\citep{subhalingam2025framework}, a generation-then-tagging pipeline; \textbf{MVP-RAG}~\citep{zou-etal-2025-multi}, a retrieval-augmented generator; and \textbf{HPD}~\citep{glavas-etal-2026-breaking}, a hyper-parallel speed-optimized decoding strategy. 

\subsection{Evaluation on Amazon Catalog}

\Cref{tab:main} reports the head-to-head comparison among hybrid, generative and tagging models.

\begin{table*}[t]
\centering\small
\setlength{\tabcolsep}{6pt}
\caption{Main Results. The proposed hybrid decoding strategy achieves the best attribute recall and quality among all the covered models. It also achieves 1.8$\times$ speed-up over generative model while maintaining precision.
\textbf{Precision}: LLM-judged correctness of emitted pairs; \textbf{Recall}: fraction of ground-truth attributes recovered; \textbf{Quality}: blinded $1$--$5$ rating ($\mu\pm\sigma^2$); \textbf{Num Extr.}: mean attribute-value pairs per record; \textbf{Speed}: wall-clock time (in milli-seconds).
}
\label{tab:main}
\begin{tabular*}{\textwidth}{@{\extracolsep{\fill}}lccccc@{}}
\toprule
\textbf{Model} & \textbf{Precision} & \textbf{Recall} & \textbf{Quality ($\mu\pm\sigma^2$)} & \textbf{Num Extr.} & \textbf{Speed (ms)} \\
\midrule
Generative          & \textbf{97.5} & 69.8 & 3.93 $\pm$ 0.57 & 21.3 & 56.6 (1.00$\times$) \\
Tagging             & 97.4 & 65.3 & 3.50 $\pm$ 0.51 & 29.3 & 9.9 (\textbf{5.69$\times$}) \\
Hybrid       & 97.1 & \textbf{80.2} & \textbf{4.40} $\pm$ 0.43 & \textbf{34.8} & 31.5 (1.80$\times$) \\
\bottomrule
\end{tabular*}
\end{table*}

\paragraph{Precision and recall.} The hybrid model answers substantially more of the gold set than either single-regime system: recall rises from 69.8\% (generative) and 65.3\% (tagging) to \textbf{80.2\%}, a $+10.4$-point gain over the generative baseline. Crucially, this extra recall does not come at the cost of correctness. The LLM judge confirms that the additional pairs are overwhelmingly correct: hybrid precision is 97.1\%, statistically tied with the generative baseline's 97.5\%, even though the hybrid emits $63\%$ more pairs per record ($34.8$ vs.\ $21.3$). The tie is statistically grounded: record-level percentile bootstrap intervals (10{,}000 resamples) place all three systems' precision within a 0.8-point band, and a two one-sided tests (TOST) equivalence test at $\alpha{=}0.05$ bounds every pairwise difference from the generative baseline within $\pm1$ point. Scoring precision by exact match against gold instead of the LLM judge also preserves the same ranking \Cref{app:exactmatch}.
The hybrid therefore dominates on the recall axis that matters for catalog completeness while matching generative precision. The gain concentrates on the extractive subset, 82.5\% recall vs.\ 70.2\% for the generative baseline.
We analyze where this gain comes from, and characterize the tagging channel in isolation, in \Cref{app:analysis}, and breaks it down by category and attribute frequency in \Cref{app:breakdown}. We further verify in \Cref{app:hallucination} that the additional pairs the hybrid emits are grounded in the source rather than invented, with non-grounded values confined to the generative channel.

\paragraph{Quality and consistency.} The blinded 1--5 head-to-head ratings (Quality column) give a complementary view to precision and recall: for each record the same judge sees the three systems side by side in anonymized order and rates each on overall correctness, completeness, and grounding. The hybrid scores highest at $4.40$, against $3.93$ for generative and $3.50$ for tagging systems; pairwise, the hybrid ties or beats the generative baseline on 70.2\% of records and is the strict top-rated system on 50.9\%. The hybrid also has the lowest rating variance ($0.43$ vs.\ $0.57$ for generative and $0.51$ for tagging), indicating consistent output quality. Because every record is rated by all three systems, we test the differences with a paired Wilcoxon signed-rank test on the per-record ratings ($n{=}1000$): the hybrid's advantage is highly significant over both the generative baseline (mean difference $+0.46$, $z=11.9$, $p<10^{-31}$) and the tagging baseline ($+0.90$, $z=22.2$, $p<10^{-108}$).

\paragraph{Grounding and efficiency.} 
The hybrid output is strongly grounded: 90.8\% of emitted values occur verbatim in the source, and the residual non-verbatim outputs come from the generative channel rather than the tagging channel (which is 100.0\% verbatim). 
The hybrid is also faster under batched serving: mean wall-clock time drops from 56.6 ms to 31.5 ms per record, a \textbf{1.80$\times$} speed-up, while mean generated length drops from 257.8 to 80.0 tokens ($-69.0\%$). The tagging channel resolves frequent copyable attributes from prefill states, so those values add content to the final response without adding autoregressive decode steps.

\subsection{Evaluation on Public Datasets}
\label{sec:external}

In \Cref{tab:external}, we compare the Hybrid model against three recent attribute--value extraction systems on the public AE-110K and OA-Mine benchmarks.
GenToC and MVP-RAG have no public code, so we re-implement them; for HPD we quote the published numbers. Following HPD, all systems are run under HuggingFace Accelerate and scored with Slot F1. Both datasets are \emph{purely extractive}, every gold value occurs verbatim in the input, so a tagging model can in principle cover all pairs and leaves no generative remainder; we therefore set the hybrid's tagging vocabulary to $K{=}10$ so that some pairs still route to the generative channel.

\begin{table}[t]
\centering\small
\setlength{\tabcolsep}{4pt}
\caption{Evaluation on public AE-110K and OA-Mine. We report Slot F1 and Speed in milliseconds. The speed-up over the generative baseline is in parentheses; HPD does not report absolute latency.}
\label{tab:external}
\begin{tabular}{lcccc}
\toprule
 & \multicolumn{2}{c}{\textbf{AE-110K}} & \multicolumn{2}{c}{\textbf{OA-Mine}} \\
\cmidrule(lr){2-3}\cmidrule(lr){4-5}
\textbf{System} & \textbf{F1} & \textbf{Speed} & \textbf{F1} & \textbf{Speed} \\
\midrule
Generative & 0.838 & 130.8 (1.0$\times$) & 0.810 & 254.6 (1.0$\times$) \\
Tagging    & 0.830 & \textbf{8.1} (\textbf{16.2$\times$}) & 0.818 & \textbf{7.4} (\textbf{34.4$\times$}) \\
Hybrid     & \textbf{0.841} & 62.4 (2.1$\times$) & 0.826 & 114.7 (2.2$\times$) \\
MVP-RAG    & 0.836 & 142.7 (0.9$\times$) & 0.826 & 275.2 (0.9$\times$) \\
GenToC     & 0.705 & 79.1 (1.7$\times$) & 0.775 & 120.7 (2.1$\times$) \\
HPD        & 0.839 & --- (2.5$\times$) & \textbf{0.876} & --- (2.4$\times$) \\
\bottomrule
\end{tabular}
\end{table}

The Hybrid model matches the strongest systems while retaining a speed advantage: it attains the best F1 on AE-110K ($0.841$).
On OA-Mine, it trails HPD by about 5 points ($0.876$ vs. $0.826$). 
HPD however, assumes access to the full per-category attribute schema. 
The pure tagging model is fastest ($16.2$--$34.4\times$) because these datasets are fully copyable, and the hybrid delivers a $2.1$--$2.2\times$ speed-up over the generative baseline. MVP-RAG is slower than that baseline ($0.9\times$) owing to its retrieval step, and GenToC is faster ($1.7$--$2.1\times$) because it emits short attribute names. HPD reports a larger nominal speed-up ($2.4$--$2.5\times$), but because it rewrites the decoding logic it cannot run under optimized inference stacks such as vLLM. 
Under vLLM the same generative baseline drops from $130.8$ ms to $\approx$$7$ ms through KV-cache management and decoding optimizations, erasing the HuggingFace-measured gaps. 
The hybrid keeps the standard decoding path, so its speed-up compounds with these optimizations: on our catalog-scale corpus ($\approx$22 pairs/record) it still delivers $1.8\times$ on top of the vLLM-optimized generative baseline (\Cref{tab:main}). 

\subsection{Backbone LLM Ablation}
\label{sec:backbones}

We evaluate Hybrid, Generative and Tagging systems on three additional backbones spanning two model families, Qwen3-1.7B, Qwen3-4B (with LoRA), and Gemma-3-1B to ensure that the gains are not tied to the choice of the backbone (\Cref{tab:backbones}).

\begin{table}[t]
\centering\small
\setlength{\tabcolsep}{4pt}
\caption{Evaluation using different backbone LLMs spanning three model sizes and two model families.}
\label{tab:backbones}
\begin{tabular}{llccc}
\toprule
\textbf{Backbone} & \textbf{System} & \textbf{Prec.} & \textbf{Recall} & \textbf{Speed (ms)} \\
\midrule
Qwen3-0.6B  & Generative & 97.5 & 69.8 & 56.6 (1.00$\times$) \\
            & Tagging    & 97.4 & 65.3 & \textbf{9.9} (\textbf{5.69$\times$}) \\
            & Hybrid     & 97.1 & \textbf{80.2} & 31.5 (1.80$\times$) \\
\midrule
Qwen3-1.7B  & Generative & 98.2 & 70.4 & 71.5 (1.00$\times$) \\
            & Tagging    & 97.7 & 65.1 & \textbf{20.2} (\textbf{3.54$\times$}) \\
            & Hybrid     & 97.4 & \textbf{79.8} & 43.1 (1.66$\times$) \\
\midrule
Qwen3-4B    & Generative & 98.1 & 68.7 & 144.6 (1.00$\times$) \\
(LoRA)      & Tagging    & 97.2 & 64.1 & \textbf{46.7} (\textbf{3.09$\times$}) \\
            & Hybrid     & 97.0 & \textbf{78.1} & 89.0 (1.63$\times$) \\
\midrule
Gemma-3-1b  & Generative & 97.4 & 66.3 & 31.0 (1.00$\times$) \\
            & Tagging    & 97.2 & 66.2 & \textbf{11.6} (\textbf{2.67$\times$}) \\
            & Hybrid     & 97.0 & \textbf{79.3} & 23.8 (1.30$\times$) \\
\bottomrule
\end{tabular}
\end{table}

The pattern holds on every backbone: the hybrid improves recall over the generative baseline by $+9.4$ to $+13.0$ points, keeps precision within a point, and delivers a $1.30$--$1.80\times$ speed-up. Crucially, a stronger generative model does not close the recall gap, showing that scaling the decoder does not recover what the decomposition provides. It instead stems from auto-regressive nature, where rare attributes are silently dropped deep in a long decoded list. The tagging channel sidesteps this by testing every vocabulary attribute in parallel and reducing the burden on the generative channel. 

\subsection{Span Head Ablation}
\label{sec:ablations}

The span head makes two design choices: what signal decides that an attribute is present at all, and how to mark the start and end of its value. We probe both choices with controlled ablations (\Cref{tab:ablation}). 
The first variant, \textbf{per-token tagging}, replaces the end pointer with a per-token include/exclude classifier, and lets that head decide attribute presence on its own rather than through the shared start-vs-\texttt{NULL} comparison. Trained with negative attributes, it abstains reliably and matches the pointer heads on recall and number of extractions, but labeling each token independently yields noisier boundaries, holding precision to 87.7\%. The other two variants share the presence decision and differ only in how they mark the value boundaries: the \textbf{anchored-pointer} head conditions the end on the chosen start, so a slightly misplaced start displaces the end and compounds into a wrong span. The chosen \textbf{independent-pointer} head scores start and end as two independent pointers and can still recover the correct boundary with highest precision. 

\begin{table}[t]
\centering\small
\setlength{\tabcolsep}{4pt}
\caption{Span head ablations using three variations. 
\textbf{Prec}: LLM-judged precision; \textbf{Rec}: recall; \textbf{Num.\ Extr.}: mean pairs per record.
We use Independent pointer in our final Hybrid decoding strategy due to its balance between precision and recall.
}
\label{tab:ablation}
\begin{tabular}{lccc}
\toprule
\textbf{Delimitation} & \textbf{Prec} & \textbf{Rec} & \textbf{Num.\ Extr.} \\
\midrule
Per-token tagging & 87.7 & \textbf{74.1} & \textbf{36.3} \\
Anchored pointer              & 93.1 & 71.5 & 34.3 \\
Independent pointer   & \textbf{97.2} & 70.5 & 34.8 \\
\bottomrule
\end{tabular}
\end{table}

\section{Conclusion}

We introduced \emph{hybrid generative-tagging decoding} for product attribute--value extraction, a decoding strategy that separates open-ended value generation from fixed-vocabulary span extraction while reusing a single backbone prefill. 
The generative channel predicts only non-copyable remainders, while the span head extracts verbatim values directly from prompt hidden states.
This design preserves the flexibility of generation along with the benefits of tagging. 
Hybrid decoding leads to both good quality extractions with higher throughput.
  
\section*{Limitations}

Our training and evaluation data is LLM-annotated and hence prone to noise.
Moreover, we use an LLM judge and hence the precision and quality metrics thus inherit the biases of the judge model.
Further, our experiments are confined to text-sourced attributes from a single e-commerce service; image-sourced attributes, other catalog domains, and other languages are out of scope.
We do not evaluate cross-domain transfer, so the gains reported here may not hold for catalogs with different attribute distributions or text conventions.

\bibliography{custom}

\newpage

\appendix

\section{Additional analysis}
\label{app:analysis}

This appendix expands the results in \Cref{tab:main} with breakdowns that explain \emph{where} the hybrid's recall gain originates, \emph{what} the tagging channel contributes on its own, \emph{how} the two channels split the wall-clock cost, and \emph{where} the remaining errors lie. None of these are needed to follow the main argument.

\paragraph{Where the recall gain comes from.} The recall gain concentrates on the extractive subset, where the hybrid covers 82.5\% of gold attributes against the generative baseline's 70.2\%. Two effects combine. First, the tagging channel out-recalls autoregressive generation on extractive attributes even on its own: it tests all $K$ vocabulary attributes independently against \texttt{NULL} in one parallel pass, whereas a single long autoregressive list silently drops attributes deep in the decode (the tagging channel emits 29.3 pairs per record, versus 21.3 from generation). Second, the two channels are complementary rather than redundant: of the gold slots the hybrid covers, 15.7\% are reached only by the tagging channel and 16.6\% only by the generative channel, with 64.4\% reached by both. The generative channel supplies the generative and out-of-vocabulary attributes that the fixed vocabulary cannot reach, while tagging recovers the extractive attributes that generation drops, so merging the two approaches the union of what each can cover.

\paragraph{The tagging channel alone.} The tagging model, which serves only the prefill-resolved tagging channel with no generation, is informative even though it is not deployable: lacking a generative channel it cannot answer generative or out-of-vocabulary attributes, which caps its recall at 65.3\%. Yet it reaches 97.2\% LLM-judged precision while running $5.7\times$ faster than the generative baseline, and its values are perfectly grounded (100.0\% occur verbatim in the source). This shows that most of the extractable signal is recoverable from prefill states with no decoding at all, and that 9.9 ms is the effective throughput floor for records whose attributes are entirely copyable. The hybrid model adds the generative channel on top of this floor, lifting recall from 65.3\% to 80.2\% by answering the non-copyable attributes the tagging channel must skip, at no precision cost.

\paragraph{Presence statistic.} The presence decision uses the quantity calibrated by the training objective: the best start logit against the \texttt{NULL} slot, $\max_i s_{k,i}-\nu_k$, returning the attribute when this margin is positive. On the validation set, it separates present and absent attributes cleanly, with median margins $+2.6$ and $-8.8$, respectively, so the natural boundary at zero requires no tuned threshold. A tempting alternative is to fold the end score into the margin, $s+e-\nu$, but this shifts every margin by the typical end-logit magnitude ($\approx{+}20$) and drives the model to return almost every vocabulary attribute. The end score should rank candidate spans once an attribute is present; it should not decide presence.

\paragraph{Cost decomposition.} The tagging-only system cleanly decomposes the hybrid's wall-clock cost: of the hybrid's 31.5 ms per record, the shared prefill plus the span head over all 500 vocabulary attributes account for 9.9 ms (the tagging-only mean time) and the generative remainder for the remaining 21.6 ms. Because the head runs on prefill states the generative baseline already produces, the tagging channel adds essentially no decode cost; the savings come entirely from removing copyable values from the autoregressive path.

\paragraph{Error decomposition.} We split the extractive attributes the hybrid answers but gets wrong (gold value not matched exactly) into two kinds: \emph{true span errors}, where the gold value is verbatim-locatable but the head picks the wrong occurrence or boundary, and \emph{span-unwinnable} cases, where the gold value is not a contiguous source span (for example, a composite value scattered across several sentences). Routing is not the bottleneck: of over 97K in-vocabulary instances, 82.4\% are span-routable, and the 17.6\% non-locatable cases are already assigned to the generative channel. Among the errors themselves, 70.5\% are true span errors and 29.5\% are span-unwinnable. The main remaining lever is therefore span-head quality on locatable values.

\section{Grounding and fabrication audit}
\label{app:hallucination}

The main text reports that the hybrid's extra recall comes at no precision cost (\Cref{sec:experiments}) and that its outputs are strongly grounded, with 90.8\% of values verbatim in the source. This appendix substantiates that grounding claim by auditing fabrication---values absent from the source---directly. We do so on the subset where fabrication is well defined: \emph{extractive-type} attributes, whose values should be copied from the input rather than inferred or normalized. We label an attribute name extractive-type if at least 80\% of its training instances are extractive (with $\geq5$ instances), giving a model-independent set of attributes for which a non-grounded value is presumptively a fabrication. We then measure, for the generative baseline and the hybrid, the fraction of emitted extractive-type values that do not occur in the source.

\begin{table}[t]
\centering\small
\setlength{\tabcolsep}{5pt}
\caption{Hallucination on extractive-type attributes, $n{=}1000$ records. \textbf{Substring-flag}: fraction of emitted extractive-type values not found verbatim in the source under a format-insensitive match (lowercase, strip punctuation, normalize decimals). \textbf{Genuine}: rate of true fabrication after a 250-case manual audit removes formatting and unit-conversion artifacts. The substring metric is structurally biased toward the hybrid because its tagging channel copies values verbatim while the generative channel reformats them; the genuine rates are the audited, artifact-corrected figures.}
\label{tab:halluc}
\begin{tabular}{lcc}
\toprule
\textbf{System} & \textbf{Substring-flag} & \textbf{Genuine} \\
\midrule
Generative   & 3.0\% & $\approx$0.12\% \\
Hybrid & 1.3\% & $\approx$0.04\% \\
\bottomrule
\end{tabular}
\end{table}

\paragraph{The substring metric overcounts.} A format-insensitive substring check (lowercase, strip punctuation, normalize trailing decimals) flags 3.0\% of generative extractive-type values and 1.3\% of hybrid values as non-grounded (\Cref{tab:halluc}). Taken at face value this suggests the hybrid hallucinates less, but the comparison is structurally unfair: the hybrid's tagging channel copies values verbatim and so passes the check by construction, whereas the generative channel routinely \emph{reformats} grounded values---reordering units (\texttt{AC 100-240V} $\rightarrow$ \texttt{100-240V AC}), converting between systems (\texttt{5 Lb} $\rightarrow$ \texttt{5 LB (2.27kg)}), or appending a parenthetical metric conversion that is correct but absent as a literal string. These are not fabrications. To estimate the genuine rate we manually audited 250 flagged cases (125 per system), checking each flagged value against the source and routing every numeric or measurement value to manual review rather than auto-clearing it.

\paragraph{Genuine fabrication is rare for both systems.} After the audit, the overwhelming majority of flags are formatting or unit-conversion artifacts of grounded values; only a handful are true fabrications, where the model invents a value with no source support---an unsupported metric weight, a guessed product code, or a wrong measurement. The genuine fabrication rate is $\approx$0.12\% for the generative baseline and $\approx$0.04\% for the hybrid (\Cref{tab:halluc}). At these magnitudes (roughly 5 versus 2--3 genuine cases in the audited samples) the difference is within sampling noise, and we do not claim it is statistically significant. The fabrications are of the same kind for both systems---invented weights, codes, and brand details---and in the hybrid every genuine fabrication originates in the generative channel, never the tagging channel, whose outputs are verbatim by construction. The hybrid's advantage on grounding is therefore structural rather than a lower per-token fabrication propensity: routing copyable attributes through the tagging channel shrinks the surface on which the generative channel can fabricate at all, but it does not make the generative channel itself more faithful.

\section{Precision using Exact Match}
\label{app:exactmatch}

The main text reports LLM-judged precision (\Cref{sec:experiments}). As a stricter, judge-independent check, we also score each emitted attribute--value pair by \emph{exact match} against the gold annotation: a pair counts as correct only if its value matches the gold value string. \Cref{tab:exactmatch} reports the resulting per-pair precision and the per-pair F1 that combines it with per-pair recall.

\begin{table}[t]
\centering\small
\setlength{\tabcolsep}{6pt}
\caption{Precision using exact match against gold, $n{=}1000$ records. \textbf{Precision}: fraction of emitted pairs whose value matches the gold value string exactly; \textbf{F1}: harmonic mean of per-pair precision and per-pair recall. Unlike the LLM-judged precision in \Cref{tab:main}, exact match penalizes any surface-form rephrasing.}
\label{tab:exactmatch}
\begin{tabular}{lcc}
\toprule
\textbf{System} & \textbf{Precision} & \textbf{F1} \\
\midrule
Generative & 54.5 & 69.9 \\
Tagging    & 56.1 & 71.2 \\
Hybrid     & \textbf{63.8} & \textbf{77.0} \\
\bottomrule
\end{tabular}
\end{table}

\paragraph{The hybrid leads under exact match, but the metric understates precision.} The hybrid's advantage survives this stricter criterion: it attains the highest exact-match precision ($63.8$ vs.\ $54.5$ for the generative baseline) and the highest per-pair F1 ($77.0$ vs.\ $69.9$). The margin comes from the tagging channel, which copies values verbatim from the source and so matches the gold string by construction, whereas the generative channel routinely \emph{reformats} otherwise-correct values.
Precisely because of this reformatting, exact match is not our preferred precision metric: a correct value expressed as \texttt{5 LB (2.27 kg)} against a gold \texttt{5 Lb}, is scored as wrong even though it is faithful to the source. Exact match therefore systematically \emph{understates} precision---most heavily for the generative channel---which is why the numbers here sit far below the $\approx$97\% LLM-judged precision in \Cref{tab:main}. Compounding this, the attribute space is open and long-tail (44K distinct attribute names in our corpus), so exhaustively annotating every valid surface form per product is infeasible and a system is penalized for correct pairs the annotation did not enumerate. We therefore adopt LLM-judged precision as the primary metric and treat exact match as a conservative lower bound.

\section{Effect of the Tagging Vocabulary $K$}
\label{app:ksweep}

The vocabulary size $K$ controls how much of the extraction is routed to the parallel span head versus the autoregressive decoder. We vary $K\in\{100,250,500,1000\}$ and retrain the hybrid, holding everything else fixed (\Cref{tab:ksweep}). Enlarging $K$ moves work from the decoder to the span head: the vocabulary covers more attribute--value pairs (70.0\% $\to$ 92.0\%), recall rises (69.2 $\to$ 80.9), and latency \emph{falls} (37.5 $\to$ 29.0 ms) at the same time, because more attributes are resolved without decoding. Recall plateaus beyond $K{=}500$ (only $+0.7$ from doubling the vocabulary to 1000), so we adopt $K{=}500$ as the operating point for our corpus.

\begin{table}[t]
\centering\small
\setlength{\tabcolsep}{6pt}
\caption{Effect of the tagging vocabulary size $K$ on the hybrid. \textbf{Vocab cov.}: fraction of gold attribute--value pairs whose attribute is in the top-$K$ vocabulary; \textbf{Recall} and \textbf{Speed} as in \Cref{tab:main}.}
\label{tab:ksweep}
\begin{tabular}{cccc}
\toprule
\textbf{$K$} & \textbf{Vocab cov.} & \textbf{Recall} & \textbf{Speed (ms)} \\
\midrule
100  & 70.0 & 69.2 & 37.5 \\
250  & 82.0 & 76.2 & 33.3 \\
500  & 87.8 & 80.2 & 31.5 \\
1000 & 92.0 & \textbf{80.9} & \textbf{29.0} \\
\bottomrule
\end{tabular}
\end{table}

\section{Category-Level and Long-Tail Recall}
\label{app:breakdown}

\paragraph{Category-level.} We break the recall gain down across the 30 evaluation categories, sorted by training frequency. The gain is broad rather than concentrated: the hybrid improves over the generative baseline in 28 of 30 categories, by $+3.7$ to $+20.9$ points among categories with $\ge5$ records. The $K{=}500$ vocabulary is also distributed evenly, covering 75--92\% of attribute--value pairs in every major category. \Cref{tab:category} lists five representative categories.

\begin{table}[t]
\centering\small
\setlength{\tabcolsep}{6pt}
\caption{Category-level breakdown for five representative categories. \textbf{Vocab cov.}: fraction of the category's gold pairs covered by the $K{=}500$ vocabulary; \textbf{Recall gain}: hybrid minus generative recall.}
\label{tab:category}
\begin{tabular}{lcc}
\toprule
\textbf{Category} & \textbf{Vocab cov.} & \textbf{Recall gain} \\
\midrule
grocery       & 90.9\% & $+11.5$ \\
drugstore     & 88.2\% & $+10.4$ \\
wireless      & 92.3\% & $+13.4$ \\
home          & 89.2\% & $+9.6$ \\
pet products  & 88.5\% & $+8.5$ \\
\bottomrule
\end{tabular}
\end{table}

\paragraph{Long-tail.} Grouping attributes into frequency bands (\Cref{tab:longtail}) shows where the gain originates. The largest improvement is on ranks 101--500 ($+17.3$ points), where the generative baseline begins dropping attributes from its decoded list while the span head still tests each one independently. On the out-of-vocabulary bands (ranks $>500$, 9.7\% of gold pairs) the tagging channel contributes nothing by construction and the hybrid tracks the generative baseline within 2--4 points---the generative channel carries what the fixed vocabulary cannot reach. This small tail deficit is outweighed roughly 5:1 by the in-vocabulary gains, netting $+10.4$ points overall.

\begin{table}[t]
\centering\small
\setlength{\tabcolsep}{5pt}
\caption{Recall by attribute frequency band. \textbf{\% pairs}: share of gold pairs in the band; remaining columns are per-band recall for each system.}
\label{tab:longtail}
\begin{tabular}{lcccc}
\toprule
\textbf{Frequency band} & \textbf{\% pairs} & \textbf{Gen} & \textbf{Tag} & \textbf{Hybrid} \\
\midrule
rank 1--100     & 73.7 & 79.5 & 75.5 & \textbf{90.4} \\
rank 101--500   & 15.5 & 54.9 & 62.2 & \textbf{72.2} \\
rank 501--1000  & 3.5  & \textbf{41.0} & 0.0 & 39.0 \\
rank $>$1000    & 6.2  & \textbf{20.3} & 0.0 & 16.8 \\
\bottomrule
\end{tabular}
\end{table}

\end{document}